\documentclass[10pt,twocolumn,letterpaper]{article}

\usepackage{cvpr}              % To produce the CAMERA-READY version
\definecolor{cvprblue}{rgb}{0.21,0.49,0.74}
\usepackage[pagebackref,breaklinks,colorlinks,allcolors=cvprblue]{hyperref}
\usepackage{booktabs}
\usepackage{multirow}
\usepackage{graphicx}
\def\paperID{} % *** Enter the Paper ID here
\def\confName{CVPR}
\def\confYear{2026}

\title{FF3R: Feedforward Feature 3D Reconstruction from Unconstrained views}

\author{
Chaoyi Zhou$^{1,2}$,
Run Wang$^2$,
Feng Luo$^2$,
Mert D. Pesé$^2$,\\
Zhiwen Fan$^3$,
Yiqi Zhong$^{\dagger1}$,
Siyu Huang$^2$
\\
\\
$^1$Microsoft
\quad \quad 
$^2$Clemson University
\quad \quad 
$^3$Texas A\&M University\\
\\
}
\begin{document}
\twocolumn[{
\maketitle

\vspace{-1cm}
\begin{center}
\includegraphics[width=.8\linewidth]{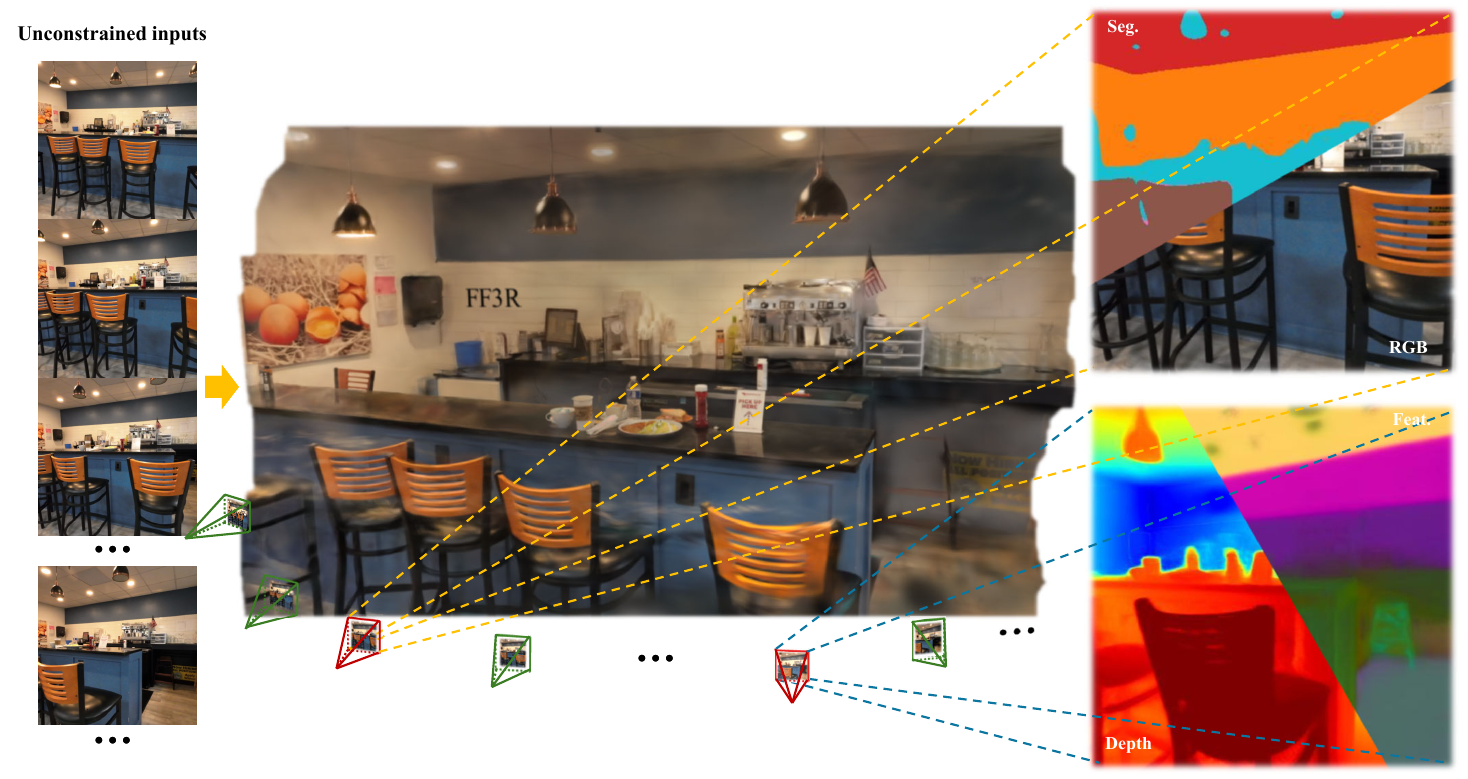}
\end{center}
\vspace{-0.5cm}
\captionsetup{type=figure}
\captionof{figure}{FF3R is the first scalable, fully self-supervised, feed-forward framework that unifies geometric and semantic reasoning from unconstrained multi-view image sequences. It achieves strong performance in both 3D reconstruction and scene-level understanding.}\label{fig:teaser}
\vspace{0.3cm}
}] 

\def\thefootnote{$\dagger$}\footnotetext{Corresponding author. The work was done during Chaoyi Zhou’s internship at Microsoft.}
\begin{abstract}

Recent advances in vision foundation models have revolutionized geometry reconstruction and semantic understanding. Yet, most of the existing approaches treat these capabilities in isolation, leading to redundant pipelines and compounded errors. 
This paper introduces FF3R, a fully annotation-free feed-forward framework that unifies geometric and semantic reasoning from unconstrained multi-view image sequences. 
Unlike previous methods, FF3R does not require camera poses, depth maps, or semantic labels, relying solely on rendering supervision for RGB and feature maps, establishing a scalable paradigm for unified 3D reasoning.
In addition, we address two critical challenges in feedforward feature reconstruction pipelines, namely global semantic inconsistency and local structural inconsistency, through two key innovations: (i) a Token-wise Fusion Module that enriches geometry tokens with semantic context via cross-attention, and (ii) a Semantic–Geometry Mutual Boosting mechanism combining geometry-guided feature warping for global consistency with semantic-aware voxelization for local coherence. Extensive experiments on ScanNet and DL3DV-10K demonstrate FF3R’s superior performance in novel-view synthesis, open-vocabulary semantic segmentation, and depth estimation, with strong generalization to in-the-wild scenarios, paving the way for embodied intelligence systems that demand both spatial and semantic understanding. Project page: \url{https://chaoyizh.github.io/ff3r_project}. 

\vspace{-0.5cm}
\end{abstract}    
\section{Introduction}
\label{sec:intro}

Vision foundation models have recently revolutionized both geometric reconstruction and semantic understanding. Geometry models~\cite{dust3r_cvpr24,wang2025continuous,wang2025vggt} replace slow optimization-based methods~\cite{sfm,mildenhall2020nerf,kerbl3Dgaussians} with scalable feed-forward systems that reconstruct 3D structures from hundreds of unconstrained images in a single pass. Meanwhile, semantic models~\cite{clip,zhai2023sigmoid,lseg,dino,oquab2023dinov2} unify recognition pipelines, achieving strong vision–language alignment and rich open-vocabulary semantics. Both systems take image sequences as input to deliver geometric or semantic understanding, two pillars for modern intelligent applications such as robotic navigation and multimodal agentic systems. However, splitting these capabilities into separate frameworks does not just add redundancy; it compounds error propagation and bloats the pipeline into a brittle, inefficient, and nearly intractable architecture. Consequently, the research community is converging on a transformative paradigm: unified systems that seamlessly fuse geometric and semantic reasoning, delivering both in a single, coherent framework.

Beyond the modality gap between geometric and semantic information, the stark differences in training strategies make building a truly unified system far from trivial. Geometry foundation models~\cite{ye2024noposplat,jiang2025anysplat} incorporate 3D Gaussian Splatting (3DGS)~\cite{wang20243d} into neural networks. The model can be trained in a self-supervised manner by leveraging the inherent stereo-geometry priors within image sequences. In contrast, 3D semantic foundation models require additional supervision, either through large-scale annotated datasets or knowledge distillation from massive pretrained vision transformers. Therefore, training a unified system must ensure access to both geometric priors and semantic labels. To tackle this challenge, existing efforts can be divided into two categories: semantic-label-dependent and semantic-label-free. In the first line of work, researchers either rely on limited existing datasets with semantic annotations~\cite{xu2025siu3r} or construct large-scale datasets featuring fine-grained 2D–3D semantic mask correspondences~\cite{li2025scenesplat,iggt2024}. While these approaches achieve impressive performance on in-domain data, their generalization remains limited due to the fixed number of classes and the time-consuming annotation works. The second line of work~\cite{fan2024largespatialmodelendtoend,tian2025uniforwardunified3dscene,sun2025uni3runified3dreconstruction} are more sustainable approach that eliminates the need for additional annotation by enabling 
annotation-free training. These methods integrate teacher-distilled learning principles from geometry foundation models~(typically leveraging photometric loss) with knowledge distillation techniques from 2D semantic foundation model training. By rendering feature maps for novel views, they enforce the joint learning of both geometric and semantic representations. 

However, existing annotation-free methods encounter two fundamental challenges, particularly when scaling to unconstrained multi-view settings (e.g., $\ge$ 32 images with minimal viewpoint variation), a scenario increasingly common in real-world applications: \textbf{(i) Global semantic inconsistency:} Semantic features from 2D foundation models (e.g., CLIP~\cite{clip}, DINO~\cite{dino,oquab2023dinov2}) lack multi-view geometric priors. Trained on single, unstructured images without 3D constraints, they fail to maintain cross-view consistency. Supervision based on these inconsistent features drives overfitting to context-specific cues, hindering coherent 3D semantic representation. \textbf{(ii) Local structural inconsistency:} Geometric models often merge neighboring Gaussian primitives to reduce memory and computation. Without semantic guidance, this merging crosses semantic boundaries, causing ambiguity and structural distortion. While semantic-aware merging could mitigate this issue, it remains largely unexplored.

Targeting to build a joint geometry and semantic prediction framework addressing the aforementioned challenges, we propose \textbf{FF3R}, a fully annotation-free feed-forward framework, as shown in Fig.~\ref{fig:teaser}. FF3R takes an unconstrained image sequence as input and is capable of both geometrical-aware novel-view synthesis and open-vocabulary semantic understanding. Two key designs in FF3R enable it to achieve the goal:

(i) \emph{Token-wise Fusion Module}: Taking the tokens output from the pretrained geometry and semantic encoders, our token-wise fusion module leverages a cross-attention mechanism that enriches geometry representations with semantic context, thereby enabling semantically aware 3D decoding.

(ii) \emph{Mutual Boosting Mechanism}: The mechanism contains two parts, where a \emph{Geometry-Guided Feature Warping loss} enforces global semantic consistency by aligning semantic features across views through geometry-based reprojection and a \emph{Semantic-Aware Voxelization} module mitigates local semantic inconsistency in dense-view scenarios by jointly weighting geometric confidence and semantic consistency, resulting in cleaner voxel features and more stable 3D geometry.
FF3R adopts a fully annotation-free training paradigm based solely on rendering supervision for RGB images and feature maps, enabling scalable learning from arbitrary in-the-wild multi-view images without requiring any explicit annotations such as camera poses, depth maps, or semantic labels.

According to the experimental results (Tab.~\ref{tab:nvs_sparse}), FF3R, equipped with these two key designs, achieves exceptional scalability in unconstrained multi-view settings. Specifically, FF3R can process over 64 images, while prior state-of-the-art methods struggle with more than 6. As the first feed-forward approach capable of handling long image sequences, it runs 180× faster than existing optimization-based methods, marking a significant leap in efficiency.

In summary, our main contributions are as follows:
\begin{itemize}[leftmargin=*]
\item Introduced \textbf{FF3R}, which to the best of our knowledge, is the first fully annotation-free
, feed-forward framework for joint geometry–semantic prediction. It enables scalable novel view synthesis and open-vocabulary scene understanding from unconstrained multi-view inputs.
\item Proposed a Semantic–Geometry Mutual Boosting mechanism, mitigating semantic and structural inconsistencies via a \emph{Geometry-Guided Feature Warping} loss for cross-view geometry alignment and a \emph{Semantic-Aware Voxelization} module for semantic-preserving aggregation.
\item Extensive experiments on ScanNet~\cite{dai2017scannet} and DL3DV-10K~\cite{ling2024dl3dv} show that FF3R achieves superior performance in novel view synthesis, open-vocabulary semantic segmentation, and depth estimation, with strong generalization to in-the-wild scenarios.
\end{itemize}

\section{Related Work}
\label{sec:related work}
Unified frameworks for geometric reconstruction and semantic understanding have rapidly become a central research focus, driven by the demand for high fidelity and minimal redundancy in modern intelligent systems. A naive solution is to couple the training and inference of two independent models, each specialized for its respective task. However, this approach is far from trivial due to stark differences in their underlying training strategies, making seamless integration a significant challenge. To address this, existing efforts generally fall into two categories: semantic-label-dependent and semantic-label-free.

\textbf{Semantic-label-dependent frameworks:}
These methods require accurate, usually human-annotated semantic label as well as explicit 3D supervision such as camera poses and depth maps as the supervision signals to train the framework. GARField \cite{garfield2024} learns a scale-conditioned 3D affinity field by lifting multi-view SAM \cite{kirillov2023segany} masks via contrastive learning.
% , enabling consistent hierarchical scene grouping and object decomposition from posed images.
SAGA~\cite{cen2023saga} and Gaussian Grouping~\cite{gaussian_grouping} extend this framework to Gaussian primitives, where each 3D Gaussian is equipped with an additional semantic parameter to model the multiview SAM masks, improving the efficiency of 3D semantic mask rendering. Targeting generalizable geometry reconstruction and semantic understanding, many works propose to equip the feed-forward geometry foundation model with pixel-aligned semantic prediction.  Most of the work~\cite{xu2025siu3r,iggt2024,li2025scenesplat,ma2025scenesplat++} relies heavily on the data annotations. SIU3R~\cite{xu2025siu3r} trains on the annotated ScanNet~\cite{dai2017scannet} dataset. SceneSplat~\cite{li2025scenesplat}, SceneSplat++ \cite{ma2025scenesplat++}, and IGGT~\cite{iggt2024} all propose different data curation pipelines based on SAM2~\cite{ravi2024sam2} and optimization-based methods~\cite{kheradmand20243d,zhou2024feature}, so that datasets with more diversity and fine-grained labels can be achieved as the supervision. Even though the model learn directly from the alignment between the pixels and semantic masks, achieving such a large-scale dataset for the foundation model pretraining is extremely resource-consuming. Moreover, most of the dataset is still restricted to the indoor domain, hindering the generalizability of the models.

\textbf{Semantic-label-free frameworks:}
They attempt to direct distill the 2D features from the foundation models such as CLIP \cite{clip,lseg}, SAM \cite{kirillov2023segany}, and DINO \cite{oquab2023dinov2,dino} into NeRF \cite{kobayashi2022distilledfeaturefields,ye2023featurenerf,lerf2023,garfield2024,lee2024open3drf} or 3DGS \cite{qin2023langsplat,gaussian_grouping,yue2024improving,cen2023saga,zuo2024fmgs,liu2024splatrajcameratrajectorygeneration,fan2024largespatialmodelendtoend,tian2025uniforwardunified3dscene,sun2025uni3runified3dreconstruction,li2025semanticsplatfeedforward3dscene,thai2025splattalk,kobayashi2022distilledfeaturefields,zhou2024feature,li2025langsplatv2highdimensional3dlanguage}. For instance, NeRF-DFF \cite{kobayashi2022distilledfeaturefields} distills the high-dimensional features into the implicit neural representation. Using the photometric loss for the image and feature rendering allows novel view rendering, open-vocabulary segmentation. Feature-3DGS \cite{zhou2024feature} and  LangSplat \cite{qin2023langsplat} extend NeRF-DFF to Gaussian primitives, allowing for more efficient training and rendering speed. However, directly optimizing high-dimensional features in 3D not only ignores the inherent global inconsistency of 2D semantic feature space, leading to unstable feature rendering and reduced 3D semantic expressiveness. Furthermore, directly optimizing the high-dimensional features in 3D space remains time-consuming. Among all the works, LSM~\cite{fan2024largespatialmodelendtoend} is the first feed-forward method that belongs to this category. It incorporates a feature Gaussian decoder to predict the pixel-aligned semantic features. However, LSM and its following works\cite{tian2025uniforwardunified3dscene,sun2025uni3runified3dreconstruction,li2025semanticsplatfeedforward3dscene} share some common issues: i) lacking in-depth interaction between geometry and semantic information; ii) overlooking the global inconsistency of the semantic feature maps; iii) not considering the redundant Gaussian primitives for longer image sequences. Therefore, its prediction quality is not ideal, and it is impossible to scale up to unconstrained multi-view images. 

In this work, we propose FF3R with two key designs: a token-wise fusion module and a Semantic–Geometry Mutual Boosting mechanism. With these two designs, FF3R effectively overcomes the challenges of unconstrained inputs and achieves state-of-the-art performance across diverse tasks and datasets.

\section{Method}
\begin{figure*}[htbp]
    \centering
    \includegraphics[width=1.0\linewidth]{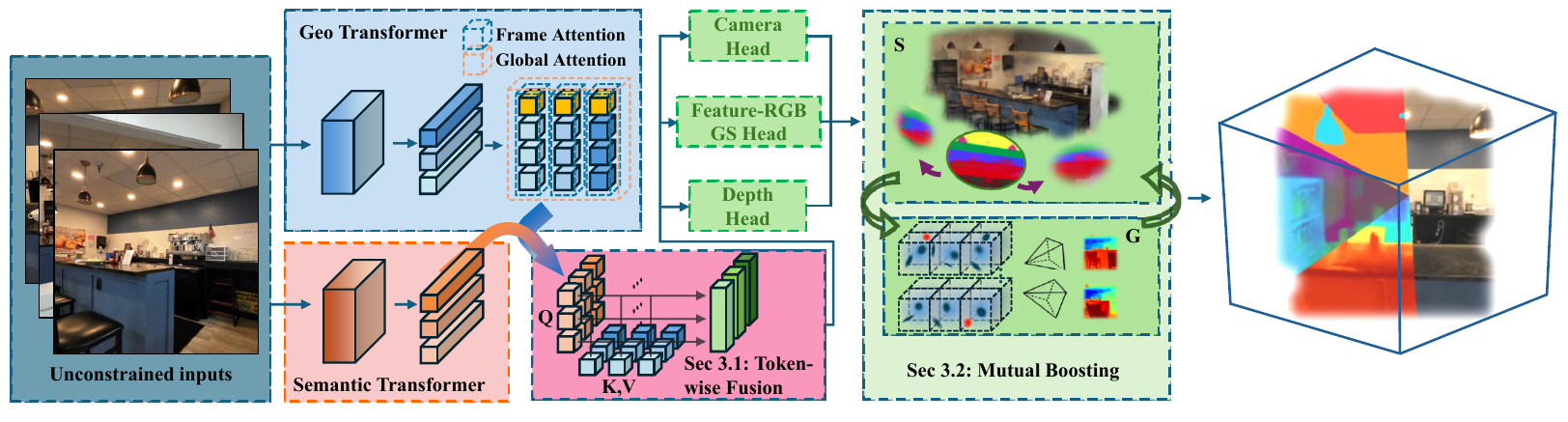}
    \vspace{-0.8cm}
    \caption{\textbf{Architecture Overview.} From unconstrained multi-view inputs, FF3R injects semantic-awareness into geometry tokens through Token-Wise Fusion, then decodes pixel-aligned features to predict feature-RGB GS, depth, and camera parameters. A Semantic–Geometry Mutual Boosting module, including Geometry-Guided Feature Warping, and Semantic-aware Voxelization, enables fully annotation-free training and yields high-quality novel view synthesis and open-vocabulary, 3D-consistent semantics.}
    \label{fig:method}
    \vspace{-0.5cm}
\end{figure*}

In this work, we propose a fully annotation-free framework that simultaneously performs 3D reconstruction and semantic understanding in a single forward pass from unconstrained multi-view images, as illustrated in Fig.~\ref{fig:method}. By leveraging a token-wise fusion module together with a Feature Gaussian decoder, our model extracts semantic-aware geometry tokens and predicts pixel-aligned geometric representations as well as semantic features, enabling simultaneous 3D reconstruction and scene-level understanding (Sec.~\ref{sec:pipeline}). To address the challenges of \textit{Global semantic inconsistency} and \textit{Local structural inconsistency}, we introduce a Semantic-Geometry Mutual Boosting mechanism (Sec.~\ref{sec:bidirection}), which encourages mutual refinement between geometry and semantics. Specifically, a \textit{Geometry-Guided Feature Warping} loss exploits geometric priors to produce 3D-consistent semantic features, while a \textit{Semantic-aware Voxelization} mitigates geometric ambiguity. Moreover, thanks to the photometric supervision and geometry distillation (Sec.~\ref{sec:ssl}), our framework can be trained in a purely annotation-free manner, eliminating the need for any explicit 3D supervision such as camera poses, depth maps, or semantic labels. Together, our proposed FF3R framework enables unified 3D reconstruction and semantic reasoning within a single forward pass, achieving high-quality and 3D-consistent open-vocabulary semantic segmentation as well as photorealistic novel view synthesis from both sparse and dense input views (Sec.~\ref{sec:exp}).

\subsection{Dense Geometry and Semantic Prediction}
\label{sec:pipeline}
 FF3R takes unconstrained multi-view images $\{ I_v \}_{v=1}^{V}$ as input, where $V \ge 2$.
Following VGGT~\cite{wang2025vggt}, we first employ DINOv2~\cite{oquab2023dinov2} to encode each image into patch tokens $\mathbf{x}_v = \{\mathbf{x}_{v,i} \mid i = 1,\ldots,N_p\}$,
where $N_p$ is the number of image patches.
To simultaneously capture semantic information, we further apply a CLIP-based segmentation encoder, LSeg~\cite{lseg}, to obtain semantic tokens $\mathbf{s}_v = \{\mathbf{s}_{v,i} \mid i = 1,\ldots,N_p\}$.
The combined image tokens $\mathbf{x}_v$, camera tokens $\mathbf{c}_v$, and register tokens $\mathbf{r}_v$ are further fed into an $L$-layer Alternating-Attention module, which enables effective information exchange both within and across frames.
Formally, the geometry tokens are obtained as $\mathbf{x}_{v}^{(L)} = f_{\mathrm{AA}}^{(L)}(\{\mathbf{x}_v, \mathbf{c}_v, \mathbf{r}_v\})$, where $f_{\mathrm{AA}}^{(L)}$ denotes the  $L$-layer Alternating-Attention module. 
\vspace{-0.3cm}
\paragraph{Token-wise Fusion} With the aim of facilitating joint geometry and semantic decoding, we propose a token-wise fusion module to enhance the semantic awareness of the geometry tokens. 
Inspired by VLM-3R~\cite{fan2025vlm3rvisionlanguagemodelsaugmented}, we employ a cross-attention mechanism where the geometry tokens serve as queries and attend to the semantic tokens, which provide both keys and values. 
This operation produces semantic-aware geometry tokens as
\begin{equation}
\mathbf{x}'_v = \mathrm{Softmax}\!\left(
   \frac{(\mathbf{x}_v^{(-1)} \mathbf{W}_Q)(\mathbf{s}_v \mathbf{W}_K)^{\!\top}}{\sqrt{d_k}}
\right) (\mathbf{s}_v \mathbf{W}_V),
\end{equation}
where $\mathbf{x}_v^{(-1)}$ denotes the final-layer geometry tokens from the Alternating-Attention module, $\mathbf{s}_v$ represents the semantic tokens obtained from LSeg, and $\mathbf{W}_Q$, $\mathbf{W}_K$, and $\mathbf{W}_V$ are learnable projection matrices. 
To better preserve the 3D-awareness and spatial consistency of the geometry representation, we perform this cross-attention only on the last layer of the geometry tokens.

\vspace{-0.3cm}
\paragraph{Feature Gaussian Decoder}
With the aim of simultaneous geometry reconstruction and semantic understanding, we adopt Feature-3DGS as the representation for our decoding. 
Following NoPoseSplat~\cite{ye2024noposplat}, we predict a set of pixel-aligned 3D Gaussian primitives 
$\mathcal{G} = \{ G_i \}_{i=1}^{N}$, 
where each primitive $G_i = (\boldsymbol{\mu}_i, \Sigma_i, \mathbf{c}_i, \alpha_i)$ \cite{kerbl3Dgaussians}
consists of a 3D mean position $\boldsymbol{\mu}_i$, 
a covariance $\Sigma_i$, 
a color feature $\mathbf{c}_i$, 
and an opacity $\alpha_i$. The color feature $\mathbf{c}_i$ is parameterized by a spherical harmonics (SH) function to model view-dependent appearance. To enable joint semantic reasoning, we further associate each Gaussian with an additional semantic feature embedding $\mathbf{f}_i$, 
resulting in an extended representation 
\[
G_i = (\boldsymbol{\mu}_i, \Sigma_i, \mathbf{c}_i, \alpha_i, \mathbf{f}_i), \quad i = 1, \ldots, N. 
\]
We adopt a DPT-based~\cite{ranftl2021visiontransformersdenseprediction} decoder, similar to VGGT \cite{wang2025vggt}, to jointly predict the dense depth map, camera parameters, and Gaussian attributes. 
Formally, given the fused geometry and semantic tokens, the decoder outputs
\[
\{\hat{D}_v,\, \hat{P}_v,\, \{\boldsymbol\Sigma_i,\, \mathbf{c}_i,\, \alpha_i,\, \mathbf{f}_i\}_{i=1}^{N}\} 
   = f_{\mathrm{DPT}}\!\left(\mathbf{x}'_v,\, \mathbf{s}_v\right),
\]
where $\hat{D}_v$ is the predicted depth map and $\hat{P}_v$ denotes the predicted camera parameters. The depth map $\hat{D}_v$ is further unprojected into the canonical 3D coordinate, serving as the center of each Gaussian primitive.
To better preserve high-frequency details from shallow layers, we introduce skip connections from early encoder stages, concatenating both the appearance features and the semantic features for each Gaussian primitive.

\subsection{Semantic-Geometry Mutual Boosting}
\label{sec:bidirection}
With the aim of eliminating the need for explicit 3D annotations such as semantic labels or camera poses, our model is supervised solely by the rendering loss computed on both rendered RGB images and feature maps. However, unlike annotated semantic masks or natural images, the semantic features inherently lack stereo-geometry priors. Consequently, direct optimization based only on the consistency loss for such inconsistent semantic features limits the training stability and generalizability to novel viewpoints. We therefore propose a \textbf{Semantic–Geometry Mutual Boosting} mechanism, which enables accurate geometry–feature alignment and promotes joint improvement of 3D reconstruction and semantic understanding. 

\begin{figure}[!t]
    \vspace{-0.3cm}
    \centering
    \includegraphics[width=0.7\linewidth]{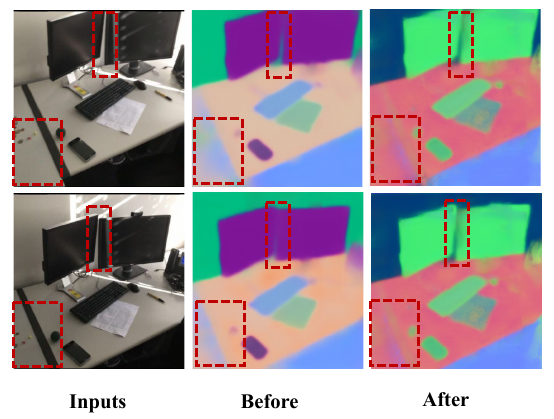}
    \vspace{-0.3cm}
    \caption{\textbf{Geometry-Guided Feature Warping.} Before warping (middle), features show inconsistency (color shifts and boundary misalignment). After warping (right), features are spatially aligned across views and better match the ground-truth semantics, yielding crisper boundaries and fewer artifacts.}
    \label{fig:geo2sem}
    \vspace{-0.5cm}
\end{figure}

\vspace{-0.4cm}
\paragraph{Geometry-Guided Feature Warping}
Recent visual foundation models such as CLIP~\cite{clip} and DINO~\cite{dino,oquab2023dinov2} are trained on unstructured web-scale image collections without explicit multiview constraints. Although these models can extract abundant semantic information from single images, their understanding of the underlying spatial structure remains limited. As shown in Fig.~\ref{fig:geo2sem}, the PCA~\cite{shlens2014tutorialprincipalcomponentanalysis} visualization of CLIP-LSeg features reveals an inherent issue: the lack of multi-view consistency. Purely optimizing these features with per-view consistency loss leads the model to overfit the context views. 

Thus, with the aim of providing the model with a multiview-consistent supervision, we propose a \textbf{Geometry-Guided Feature Warping} loss. Given two views $(I_t, I_c)$ with render feature maps $F_t, F_c$,  predicted depths $D_t, D_c$, 
intrinsics $(\mathbf{K}_t, \mathbf{K}_c)$ and relative pose $(\mathbf{R}_{t\rightarrow c}, \mathbf{T}_{t\rightarrow c})$, 
each target pixel $\mathbf{x}_t$ in $I_t$ is projected to $I_c$ via
\begin{equation}
    \mathbf{x}_c = \Pi\!\left(
        \mathbf{K}_c\left(
            \mathbf{R}_{t\rightarrow c}\,
            D_t(\mathbf{x}_t)\mathbf{K}_t^{-1}\mathbf{x}_t
            + \mathbf{T}_{t\rightarrow c}
        \right)
    \right),
\end{equation}
where $\Pi(\cdot)$ denotes perspective projection.
Using $\mathbf{x}_c$ as sampling coordinates, 
we obtain the warped feature $\mathbf{f}_{c \rightarrow t} = \text{GridSample}(F_c, \tilde{\mathbf{x}}_c)$, 
and define the cosine-similarity distance between the two feature maps as
\begin{equation}
    L_{\text{dist}}(I_t, I_c)
    =
    \frac{1}{|\Omega|}
    \sum_{\mathbf{x}_t \in \Omega}
    \mathcal{M}_{c \rightarrow t}
    \Big(
        1 -
        \frac{
            \mathbf{f}_t(\mathbf{x}_t) \mathbf{f}_{c \rightarrow t}(\mathbf{x}_t)
        }{
            \|\mathbf{f}_t(\mathbf{x}_t)\|_2
            \|\mathbf{f}_{c \rightarrow t}(\mathbf{x}_t)\|_2
        }
    \Big),
\end{equation}
where $\mathcal{M}_{c \rightarrow t}$ is the valid mask combining in-bounds and depth-consistency checks,
and $\Omega$ is the set of valid pixels.
The final bidirectional warping supervision sums over both directions:
\begin{equation}
    \mathcal{L}_{\text{warp}}
    =
    \sum_{(I_t, I_c) \in \mathcal{P}}
    \big(
        L_{\text{dist}}(I_t, I_c)
        + 
        L_{\text{dist}}(I_c, I_t)
    \big),
\end{equation}
where $\mathcal{P}$ denotes all sampled view pairs from the context views. By introducing such multi-view consistent supervision, 
our feature-based 3D Gaussian representation effectively avoids overfitting to the context views 
and learns more geometry-consistent semantic features in the shared 3D space.

% \begin{figure}[ht] % [ht] are placement specifiers (here, "here" or "top")
%     \centering % Centers the entire figure
%     \begin{subfigure}[b]{0.45\textwidth} % [b] for bottom alignment, 0.45\textwidth for width
%         \centering
%         \includegraphics[width=0.8\linewidth]{figs/mutual_1.png} % Adjust image width within subfigure
%         \caption{Geometry-Guided Feature Warping.}
%         \label{fig:geo2sem}
%     \end{subfigure}
%     \\
%     \begin{subfigure}[b]{0.45\textwidth}
%         \centering
%         \includegraphics[width=0.8\linewidth]{figs/mutual_2.png}
%         \caption{Semantic-aware Voxelization.}
%         \label{fig:sem2geo}
%     \end{subfigure}
%     \caption{Semantic–Geometry Bidirectional Interaction}
%     \label{fig:main_figure_label}
% \end{figure}
\begin{figure}[!t]
    \vspace{-0.3cm}
    \centering
    \includegraphics[width=0.7\linewidth]{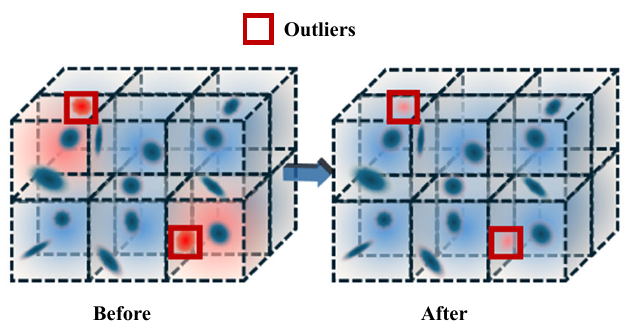}
    \vspace{-0.3cm}
    \caption{Semantic-aware Voxelization}
    \label{fig:sem2geo}
    \vspace{-0.5cm}
\end{figure}

\vspace{-0.4cm}
\paragraph{Semantic-aware Voxelization}
The pixel-aligned Feature 3DGS prediction works well for the existing method \cite{fan2024largespatialmodelendtoend}. 
However, as shown in Table~\ref{tab:nvs_sparse}, its scalability is limited when the input views become dense. 
To address this issue, AnySplat~\cite{jiang2025anysplat} introduces a Differentiable Voxelization~\cite{scaffoldgs} 
to effectively reduce redundant Gaussians by clustering pixel-aligned predictions into voxel-level representations. 
Specifically, it performs confidence-aware weighted averaging based on the predicted per-pixel confidence scores. 
While this strategy successfully compresses the number of Gaussian primitives, 
it fails to handle semantic ambiguity within a voxel. 
As illustrated in Fig.~\ref{fig:sem2geo}, an outlier Gaussian may be assigned a higher confidence 
than its surrounding points and thus dominate the voxel aggregation. 
Consequently, even though most Gaussians within the voxel are semantically consistent, the fused voxel feature becomes contaminated, leading to a distorted appearance.

To address this problem, we introduce a \textbf{Semantic-aware Voxelization} strategy 
that enforces semantic consistency within each voxel, resulting in a more coherent appearance representation. 
Following AnySplat~\cite{jiang2025anysplat}, we cluster all Gaussian centers 
$\{\boldsymbol{\mu}_g\}_{g=1}^{G}$ into a set of $S$ voxels of size $\epsilon$ 
through differentiable quantization:
\begin{equation}
\{\boldsymbol{V}_s\}_{s=1}^{S}
\;=\;
\left\lfloor
\frac{\{\boldsymbol{\mu}_g\}_{g=1}^{G}}{\epsilon}
\right\rceil,
\end{equation}
where $\boldsymbol{V}_s \in \{1,\dots,S\}$ denotes the voxel index of Gaussian $g$.
Each voxel $\boldsymbol{V}_s$ thus represents a spatial cluster 
of Gaussians within a cubic region of size $\epsilon$.

For each voxel $\boldsymbol{V}_s$, we define its semantic prototype 
$\bar{\mathbf{f}}^{sem}_s$ as the average semantic feature of all Gaussians assigned to it.
The semantic consistency of each Gaussian is then measured by the cosine distance 
to its voxel prototype:
\begin{equation}
d_g^{sem} = 1 -
\frac{\mathbf{f}^{sem}_g \cdot \bar{\mathbf{f}}^{sem}_s}{
\|\mathbf{f}^{sem}_g\|_2 \|\bar{\mathbf{f}}^{sem}_s\|_2}.
\end{equation}
The final fusion weight combines semantic consistency and geometric confidence:
\begin{equation}
w_{g \rightarrow s} =
\frac{\exp(C_g - \lambda d_g^{sem})}{
\sum_{h:V_h=s}\exp(C_h - \lambda d_h^{sem})},
\end{equation}
where $\lambda$ controls the influence of semantic distance.
Any per-Gaussian attribute $\mathbf{a}_g$ (e.g., position, color, or feature embedding) 
is then aggregated within voxel $V_s$ as:
\begin{equation}
\bar{\mathbf{a}}_s = \sum_{g:V_g=s} w_{g \rightarrow s} \, \mathbf{a}_g.
\end{equation}
This semantic–confidence joint weighting effectively suppresses semantically inconsistent outliers, 
yielding cleaner voxel features and more coherent 3D geometry.

\subsection{Learning Objective}
To ensure that the rendered RGB images from our Feature-3DGS representation align with the input context views, 
we employ a reconstruction loss that jointly minimizes pixel-wise color discrepancy and perceptual difference:
\begin{equation}
    \mathcal{L}_{\text{rgb}} =
    \| I - \hat{I} \|_1
    + \lambda_{\text{lpips}} \cdot \text{LPIPS}(I, \hat{I}),
\end{equation}
where \(I\) and \(\hat{I}\) denote the ground-truth and rendered RGB images, respectively,
and \(\lambda_{\text{lpips}}\) controls the perceptual weighting, which is set to 0.05.
This combination preserves low-frequency color fidelity while enhancing perceptual sharpness and high-frequency details.

For feature-level supervision, we enforce semantic consistency between the rendered feature map \(\hat{F}\)
and the CLIP-LSeg~\cite{lseg} feature \(F\)
using a cosine similarity loss:
\begin{equation}
    \mathcal{L}_{\text{feat}} =
    1 - 
    \frac{\hat{F} \cdot F}{
    \|\hat{F}\|_2 \, \cdot\|F\|_2},
\end{equation}
which provides open-vocabulary semantic guidance and ensures consistent feature alignment across multiple views.

To remove the need for explicit 3D annotations such as depth or camera parameters,
we adopt the distillation strategy of AnySplat~\cite{jiang2025anysplat}
using pseudo-labels generated by the pretrained VGGT~\cite{wang2025vggt} model.
The predicted camera parameters \(p_i\) and rendered depth maps \(\hat{D}_i\) 
are regularized with their corresponding pseudo ground-truths \(\tilde{p}_i\) and \(\tilde{D}_i\).
The distillation losses are defined as:
\begin{equation}
    \mathcal{L}_{\text{p}} = 
    \frac{1}{N} \sum_{i=1}^{N}
    \left\| \tilde{p}_i - p_i \right\|_{\epsilon},
    \label{eq:lp}
\end{equation}
where \(\tilde{p}_i\) denotes the pseudo pose encoding and \(\|\cdot\|_{\epsilon}\) is the Huber loss. 
We further distill the depth prediction using:
\begin{equation}
    \mathcal{L}_{\text{d}} =
    \frac{1}{N}
    \sum_{i=1}^{N}
    (\tilde{D}_i[M] - \hat{D}_i[M])^2,
    \label{eq:ld}
\end{equation}
where \(M\) is a confidence-based geometry mask selecting the top \(N\%\) most reliable pixels.

Finally, by integrating the proposed Geometry-Guided Feature Warping loss, as in Sec. \ref{sec:bidirection},
our complete training objective becomes:
\begin{equation}
\begin{aligned}
    \mathcal{L}_{\text{total}} &= \mathcal{L}_{\text{rgb}}
    + \lambda_{\text{1}}\, \mathcal{L}_{\text{feat}}
    + \lambda_{\text{2}}\, \mathcal{L}_{\text{warp}} 
    + \lambda_{\text{3}}\, \mathcal{L}_{\text{d}}
    + \lambda_{\text{4}}\, \mathcal{L}_{\text{p}},
\end{aligned}
\end{equation}
where \(\lambda_{\text{1}}\), \(\lambda_{\text{2}}\), \(\lambda_{\text{3}}\), and \(\lambda_{\text{4}}\) are set to 0.1, 0.1, 1.0, and 10.0 respectively.
This unified formulation allows our model to be trained in a fully annotation-free manner without any explicit 3D ground-truth annotations, favoring strong generalization and scalability 
to arbitrary in-the-wild multi-view inputs without hand-crafted supervision.

\label{sec:ssl}
\section{Experiments}
\label{sec:exp}

\begin{table*}[t!]
\caption{\textbf{Quantitative Results on Novel View Synthesis and Semantic Segmentation.} We evaluate sparse-view NVS and segmentation on ScanNet~\cite{dai2017scannet}, and dense-view NVS and segmentation on DL3DV-10K~\cite{ling2024dl3dv}, reporting 3D reconstruction time, image-quality metrics, and segmentation metrics.}
\vspace{-0.3cm}
\centering
\renewcommand{\arraystretch}{1.15}
\setlength{\tabcolsep}{6pt}
\resizebox{0.9\linewidth}{!}{
\begin{tabular}{l|cccccc|cccccc|cccccc}
\toprule
\multicolumn{1}{c|}{\multirow{2}{*}{Sparse}} & \multicolumn{6}{c|}{2 Views} & \multicolumn{6}{c|}{6 Views} & \multicolumn{6}{c}{16 Views} \\
\cmidrule(lr){2-7} \cmidrule(lr){8-13} \cmidrule(lr){14-19}
 & PSNR$\uparrow$ & SSIM$\uparrow$ & LPIPS$\downarrow$ & mIoU$\uparrow$ &Acc.$\uparrow$ &Time(s)$\downarrow$ 
 & PSNR$\uparrow$ & SSIM$\uparrow$ & LPIPS$\downarrow$ & mIoU$\uparrow$ &Acc.$\uparrow$ &Time(s)$\downarrow$ 
 & PSNR$\uparrow$ & SSIM$\uparrow$ & LPIPS$\downarrow$ & mIoU$\uparrow$ &Acc.$\uparrow$ &Time(s)$\downarrow$ \\
\midrule
\multicolumn{13}{l}{\textit{ScanNet ~\cite{dai2017scannet} Dataset}} \\
LSeg~\cite{lseg} & -- & -- & -- & \textbf{0.543} & \textbf{0.810} & -- & -- & -- & -- & \textbf{0.551} & \textbf{0.804} & -- & -- & -- & -- & \textbf{0.540} & \textbf{0.800} & -- \\
Feature-3DGS~\cite{zhou2024feature} & \underline{17.54} & \underline{0.673} & {0.437} & {0.332} & {0.662} & {18min} & \underline{18.34} & \underline{0.695} & \underline{0.430} & {0.330} & {0.661} & {18min} & \underline{19.12} & \underline{0.713} & \underline{0.419} & {0.349} & {0.648} & {18min} \\
LSM~\cite{fan2024largespatialmodelendtoend} & {14.95} & {0.589} & \underline{0.401} & {0.424} & {0.739} & \textbf{0.408s} & {14.50} & {0.580} & {0.438} & {0.392} & {0.708} & 16.3s & \textit{OOM} & \textit{OOM} & \textit{OOM} & \textit{OOM} & \textit{OOM} & \textit{OOM} \\
Ours & \textbf{22.70} & \textbf{0.787} & \textbf{0.285} & \underline{0.486} & \underline{0.754} & {0.884s} & \textbf{22.19} & \textbf{0.769} & \textbf{0.295} & \underline{0.500} & \underline{0.751} & \textbf{1.2s} & \textbf{22.58} & \textbf{0.790} & \textbf{0.293} & \underline{0.492} & \underline{0.739} & \textbf{2.2s} \\
\midrule
\multicolumn{1}{c|}{\multirow{2}{*}{Dense}} & \multicolumn{6}{c|}{32 Views} & \multicolumn{6}{c|}{48 Views} & \multicolumn{6}{c}{64 Views} \\
\cmidrule(lr){2-7} \cmidrule(lr){8-13} \cmidrule(lr){14-19}
 & PSNR$\uparrow$ & SSIM$\uparrow$ & LPIPS$\downarrow$ & mIoU$\uparrow$ &Acc.$\uparrow$ &Time(s)$\downarrow$ 
 & PSNR$\uparrow$ & SSIM$\uparrow$ & LPIPS$\downarrow$ & mIoU$\uparrow$ &Acc.$\uparrow$ &Time(s)$\downarrow$ 
 & PSNR$\uparrow$ & SSIM$\uparrow$ & LPIPS$\downarrow$ & mIoU$\uparrow$ &Acc.$\uparrow$ &Time(s)$\downarrow$ \\
\midrule
\multicolumn{13}{l}{\textit{DL3DV-10K ~\cite{ling2024dl3dv} Dataset}} \\
% 3D-GS~\cite{kerbl3Dgaussians} & {} & {} & {} & -- & -- & {} & {} & {} & {} & -- & -- & {} & {} & {} & {} & -- & -- & {} \\
% Mip-Splatting~\cite{Yu2024MipSplatting} & {} & {} & {} & -- & -- & {} & {} & {} & {} & -- & -- & {} & {} & {} & {} & -- & -- & {} \\
Feature-3DGS~\cite{zhou2024feature} & {17.42} & \textbf{0.588} & \underline{0.376} & \underline{0.196} & \underline{0.494} & {30min} & {18.52} & \textbf{0.611} & {0.448} & {0.225} & {0.516} & {30min} & {18.48} & \textbf{0.618} & {0.424} & \underline{0.230} & \underline{0.508} & {30min} \\
AnySplat~\cite{jiang2025anysplat} & \underline{18.02} & {0.561} & \textbf{0.375} & -- & -- & \textbf{1.4s} & \underline{19.18} & {0.603} & \textbf{0.363} & -- & -- & \textbf{2.7s} & \textbf{19.58} & \underline{0.609} & \textbf{0.342} & -- & -- & \textbf{4.1s} \\
Ours & \textbf{18.56} & \underline{0.577} & {0.377} & \textbf{0.484} & \textbf{0.800} & \underline{3.1s} & \textbf{19.31} & \underline{0.606} & \underline{0.377} & \textbf{0.514} & \textbf{0.836} & \underline{6.1s} & \underline{19.31} & {0.608} & \underline{0.365} & \textbf{0.521} & \textbf{0.821} & \underline{9.3s} \\

\bottomrule
\end{tabular}}
\label{tab:nvs_sparse}

\vspace{-0.3cm}
\end{table*}

\vspace{-0.1cm}
\subsection{Experimental Setup}
% \vspace{0.05cm}
\textbf{Implementation Details:} 
We train our model on DL3DV-10K using multi-view RGB supervision only, without any 3D annotations. 
More implementation details are provided in the appendix.

% \vspace{-0.4cm}
\textbf{Baselines:}
We compare our framework with several representative baselines. \textbf{FF3R} serves as a unified framework for 3D reconstruction and geometric understanding. 
To evaluate the generalization ability of our pipeline with respect to the number and distribution of input views, we divide our experiments into two settings: sparse-view and dense-view. 
Among all baselines, the most relevant ones are Feature 3DGS~\cite{zhou2024feature} and LSM~\cite{fan2024largespatialmodelendtoend}. 
Since the input of LSM is limited to two views, following the post-optimization strategy of DUSt3R~\cite{dust3r_cvpr24}, we perform inference in a pairwise manner when more than two views are available, and merge the results based on the overlapping cameras to obtain the final prediction. 
We also include LSeg~\cite{lseg} as a purely 2D-based baseline for scene-level semantic understanding.
In the dense-view setting, since no comparable baseline can jointly predict geometry and semantic information with the unconstrained inputs up to 64 views, we adopt AnySplat~\cite{jiang2025anysplat} as the current state-of-the-art feed-forward method for novel view synthesis. 

% \vspace{-0.4cm}
\textbf{Metrics:}
For novel view synthesis, we use Peak Signal-to-Noise Ratio (PSNR) \cite{fardo2016formalevaluationpsnrquality}, Structural Similarity Index (SSIM) \cite{nilsson2020understandingssim}, and Learned Perceptual Image Patch Similarity (LPIPS) \cite{zhang2018perceptual}. 
For open-vocabulary semantic segmentation, we adopt mean Intersection-over-Union (mIoU) and pixel-wise Accuracy. 
For depth consistency, we report the Absolute Relative Error (Rel) and Inlier Ratio ($\tau$) with a threshold of $1.03$\cite{fan2024largespatialmodelendtoend}.

\vspace{-0.1cm}
\subsection{Experiment Results}
\vspace{-0.1cm}
\begin{figure*}[htbp]
    \centering
    \includegraphics[width=.9\linewidth]{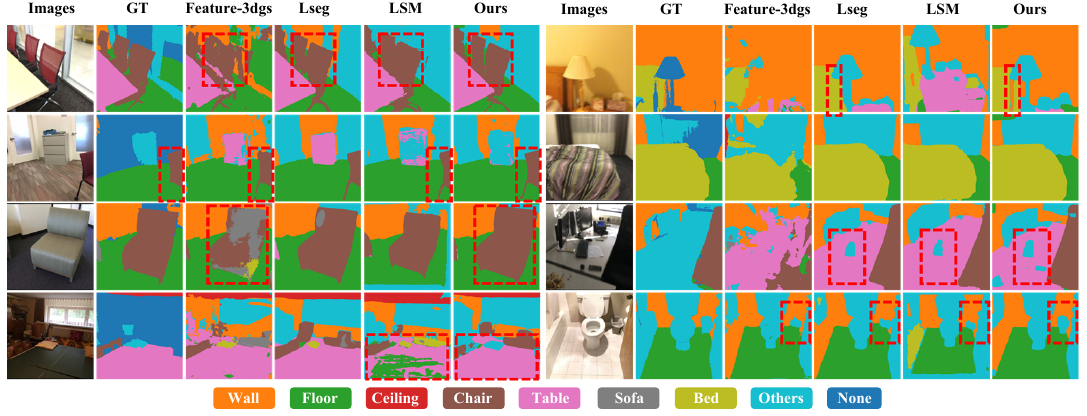}
    \vspace{-0.3cm}
    \caption{\textbf{Language-based 3D Segmentation Comparison.} Qualitative results across eight scenes from the ScanNet~\cite{dai2017scannet} dataset using varying numbers of unconstrained input views. Our FF3R produces sharper boundaries, fewer artifacts, and stronger cross-view consistency than LSM~\cite{fan2024largespatialmodelendtoend}, Feature-3DGS~\cite{zhou2024feature}, and CLIP-LSeg~\cite{lseg}, demonstrating effective fusion of semantic information and geometric structure into a coherent 3D feature field.}

    \label{fig:seg_eval}
    % \vspace{-0.3cm}
\end{figure*}

\begin{figure*}[htbp]
    \centering
    \includegraphics[width=1\linewidth]{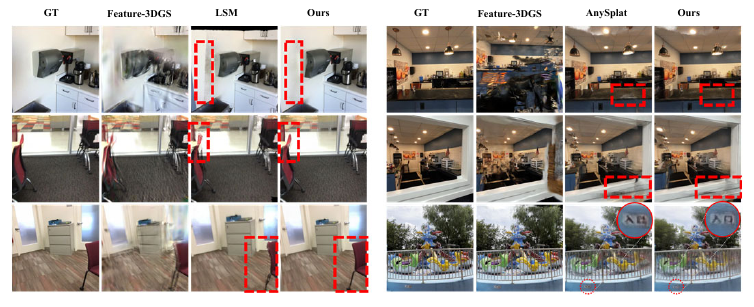}
    \vspace{-0.8cm}
    \caption{\textbf{Novel View Synthesis Comparison.} We compare results under sparse and dense view settings on the ScanNet~\cite{dai2017scannet} and DL3DV-10K~\cite{ling2024dl3dv} datasets using unconstrained inputs. FF3R consistently outperforms all baselines, achieving sharper details and higher visual fidelity across both sparse and dense scenarios.}

    \label{fig:nvs_eval}
    \vspace{-0.5cm}
\end{figure*}

\textbf{Open-Vocabulary Semantic 3D Segmentation:}
As shown in Table~\ref{tab:nvs_sparse}, FF3R largely outperforms baseline models. 
Although LSM~\cite{fan2024largespatialmodelendtoend} is capable of jointly predicting geometry and semantics without using camera poses, it cannot scale to denser-view inputs (e.g., 16 views) due to the redundant Gaussian primitives. 
Feature-3DGS~\cite{zhou2024feature}, while free from input-view limitations, requires time-consuming per-scene optimization, resulting in poor generalization ability. 
Moreover, it relies on Structure-from-Motion (SfM) results as inputs, which increases both the potential for error accumulation and the overall optimization time. 
As a purely 2D-based method, LSeg~\cite{lseg} fails to achieve spatially consistent understanding in 3D space. 
In contrast, our semantic-aware voxelization prevents linear memory growth while preserving feature quality, allowing FF3R to scale seamlessly from 2 to 64 unconstrained input views without relying on camera poses or post-optimization. 
As illustrated in Fig.~\ref{fig:seg_eval}, FF3R preserves fine-grained semantic details, particularly around object boundaries, where the proposed geometry-guided feature warping effectively incorporates 3D awareness and sharpens semantic consistency across views. Thanks to our fully annotation-free framework, FF3R demonstrates strong generalizabilities on ScanNet~\cite{dai2017scannet} and ScanNet++~\cite{yeshwanth2023scannet++}, which share similar distributions with the evaluation data but have never been seen by the model. 
This capability emerges from training on large-scale unannotated data~\cite{ling2024dl3dv}, enabling our model to perform robustly across diverse scenarios.

% \vspace{-0.4cm}
\textbf{Novel View Synthesis:}
As shown in Table~\ref{tab:nvs_sparse}, FF3R achieves high-quality novel-view rendering from sparse to dense inputs within a unified framework of simultaneous geometry reconstruction and semantic understanding. 
Although LSM~\cite{fan2024largespatialmodelendtoend} can be extended to multi-view settings through post-optimization, the lack of control over redundant Gaussian primitives prevents it from scaling to unconstrained inputs. Feature-3DGS~\cite{zhou2024feature}, as an optimization-based method, tends to overfit the context views under sparse inputs, resulting in severe distortions and degraded quality in novel views. 
While its performance improves with more input views, directly optimizing high-dimensional features in 3D space leads to a significant increase in per-scene optimization time. 
Benefiting from our semantic-aware voxelization, FF3R effectively preserves compact Gaussian representations under dense-view inputs, achieving results comparable to the state-of-the-art feed-forward method AnySplat~\cite{jiang2025anysplat}. 
As illustrated in Fig.~\ref{fig:nvs_eval}, our approach maintains local semantic consistency in challenging regions such as object boundaries and weak-texture areas, resulting in more coherent appearance reconstruction.

\begin{table}[!t]
\centering
\caption{Comparison of depth consistency under different input views. 
We report Absolute Relative Error (Rel$\downarrow$) and Inlier Ratio ($\tau\uparrow$) with a threshold of $1.03$.}
\vspace{-0.2cm}
\setlength{\tabcolsep}{6pt}
\renewcommand{\arraystretch}{1.15}
\resizebox{0.55\columnwidth}{!}{
\begin{tabular}{lcccc}
\toprule
\multirow{2}{*}{Method} & \multicolumn{2}{c}{2 Views} & \multicolumn{2}{c}{6 Views} \\
\cmidrule(lr){2-3} \cmidrule(lr){4-5}
 & Rel$\downarrow$ & $\tau\uparrow$ & Rel$\downarrow$ & $\tau\uparrow$ \\
\midrule
LSM~\cite{fan2024largespatialmodelendtoend} & 7.36 & 46.24 & 8.38 & 38.65 \\
FF3R (Ours) & \textbf{3.99} & \textbf{67.99} & \textbf{3.36} & \textbf{71.10} \\
\bottomrule
\end{tabular}
}
\label{tab:depth_consistency}
\vspace{-0.5cm}
\end{table}

\textbf{Multi-View Geometry Consistency:}
As shown in Table~\ref{tab:depth_consistency}, when scaling up the number of input views, 
LSM suffers from accumulated geometric errors introduced by repeated post-optimization steps, 
leading to degraded depth consistency. 
In contrast, our method shows further improvement as the number of views increases, 
demonstrating the effectiveness of the semantic-aware voxelization. 
By ensuring semantic consistency during the merging process, 
the richer semantic priors contribute to more stable and coherent geometric representations.

% \begin{table*}[htbp]
% \vspace{-0.2cm}
% \centering
% \caption{Ablation study of different components in FF3R. }
% \vspace{-0.2cm}
% \setlength{\tabcolsep}{6pt}
% \renewcommand{\arraystretch}{1.15}
% \resizebox{0.55\linewidth}{!}{
% \begin{tabular}{lccccccccc}
% \toprule
% Method & Token-wise Fusion & G$\rightarrow$S & S$\rightarrow$G & PSNR$\uparrow$ & SSIM$\uparrow$ & LPIPS$\downarrow$ & mIoU$\uparrow$ & Acc.$\uparrow$ \\
% \midrule
% Base &  &  &  & 17.85 & 0.699 & 0.380 & 0.411 & 0.684 \\
% + Token-wise Fusion & \checkmark &  &  & 18.70 & 0.721 & 0.359 & 0.440 & 0.721 \\
% + G$\rightarrow$S & \checkmark & \checkmark &  & 18.61 & 0.705 & 0.373 & 0.450 & 0.722 \\
% + S$\rightarrow$G & \checkmark & \checkmark & \checkmark & {19.12} & {0.718} & {0.357} & {0.449} & {0.725} \\
% \bottomrule
% \end{tabular}
% }
% \label{tab:ablation}
% \vspace{-0.5cm}
% \end{table*}

\vspace{-0.2cm}

\begin{table}[htbp]
\vspace{-0.2cm}
\centering
\caption{Ablation study of different components in FF3R. }
\vspace{-0.2cm}
\setlength{\tabcolsep}{6pt}
\renewcommand{\arraystretch}{1.15}
\resizebox{1\linewidth}{!}{
\begin{tabular}{lccccccccc}
\toprule
Method & TW Fusion & G$\rightarrow$S & S$\rightarrow$G & PSNR$\uparrow$ & SSIM$\uparrow$ & LPIPS$\downarrow$ & mIoU$\uparrow$ & Acc.$\uparrow$ \\
\midrule
Base &  &  &  & 17.85 & 0.699 & 0.380 & 0.411 & 0.684 \\
+ TW Fusion & \checkmark &  &  & 18.70 & 0.721 & 0.359 & 0.440 & 0.721 \\
+ G$\rightarrow$S & \checkmark & \checkmark &  & 18.61 & 0.705 & 0.373 & 0.450 & 0.722 \\
+ S$\rightarrow$G & \checkmark & \checkmark & \checkmark & {19.12} & {0.718} & {0.357} & {0.449} & {0.725} \\
\bottomrule
\end{tabular}
}
\label{tab:ablation}
\vspace{-0.5cm}
\end{table}

\vspace{-0.1cm}

\subsection{Ablation Study}
\vspace{-0.2cm}
As shown in Table~\ref{tab:ablation}, the base model relies solely on the Feature Gaussian Decoder for reconstruction, without semantic guidance.
Adding the token-wise fusion module~(TW Fusion) introduces semantic awareness to the geometry branch, enriching feature representations for decoding.
The G→S module further injects 3D geometric priors into semantic features, enabling spatially consistent representations and producing sharper semantic masks in 3D space. Finally, incorporating the S→G module through semantic-aware voxelization enforces a more compact Gaussian representation, ensuring fine-grained geometric structures with improved semantic consistency—ultimately leading to higher-quality appearance reconstruction. This progressive design forms a tree-structured bidirectional interaction, where semantic and geometric cues continuously refine each other, demonstrating the necessity and complementarity of all core components in our framework.

\vspace{-0.2cm}
\section{Conclusion}
\vspace{-0.1cm}
\label{sec:conclusion}
We have presented FF3R, a fully 
annotation-free feed-forward framework that unifies geometry reconstruction and semantic understanding from unconstrained multi-view images. 
By integrating a token-wise fusion module and a semantic–geometry mutual boosting mechanism, FF3R effectively bridges the gap between geometric and semantic reasoning, 
enabling high-quality novel-view synthesis, open-vocabulary semantic segmentation, and depth estimation without requiring camera poses, depth maps, or semantic labels. 
Extensive experiments on ScanNet\cite{dai2017scannet} and DL3DV-10K\cite{ling2024dl3dv} demonstrate the superior scalability and generalization of FF3R across both sparse and dense-view settings. 
We believe this work takes an important step toward large-scale, annotation-free 3D scene understanding, paving the way for next-generation embodied AI systems that require unified geometric and semantic reasoning.

{
    \small
    \bibliographystyle{ieeenat_fullname}
    \bibliography{main}
}
\appendix
\clearpage
\appendix

\newpage
\section*{Appendix}

\noindent In the Appendix, we provide the following: 
\begin{itemize}
    \item comprehensive implementation details in Section~\ref{sec:app_impl}
    \item additional experiments, results, and discussions in Section~\ref{sec:app_exp}
\end{itemize}
\section{Implementation Details}
\label{sec:app_impl}
\paragraph{Training Setup}
We train FF3R on a subset of 6,500 scenes sampled from the DL3DV-10K~\cite{ling2024dl3dv} dataset. 
No additional 3D annotations such as depth, camera poses, or semantic labels are required; 
only multi-view RGB images are used as the supervision signal. 
We initialize the geometry transformer and depth DPT head 
with pretrained weights from VGGT~\cite{wang2025vggt}, 
while all other modules are randomly initialized. 
The alternating-attention blocks are unfrozen to adapt to our downstream unified decoder structure, and we use a fixed CLIP-LSeg~\cite{lseg} as the semantic transformer.
During training, each input image is set to $448\times448$, 
and each iteration randomly samples one scene, 
from which a subset of context views (up to 16 views) is further selected. 
The model is optimized using AdamW~\cite{loshchilov2019decoupledweightdecayregularization} with a cosine learning rate scheduler, 
a peak learning rate of $2\times10^{-4}$, and a warm-up phase of 1K iterations. 
Training is performed on 8 NVIDIA A100 GPUs for two days.

\paragraph{Training View Sampling Strategy}
To enhance the robustness of our model, careful design of the training view sampling strategy is crucial. Following Dust3r\cite{dust3r_cvpr24} and VGGT \cite{wang2025vggt}, we adopt a sequential sampling approach for DL3DV \cite{ling2024dl3dv}. 
Specifically, we first randomly determine the temporal gap between the first and last frames. Within this interval, additional frames are randomly sampled to ensure that the total number of input views does not exceed 16. Since our framework imposes no requirement on temporal order, the sampled views are shuffled at each iteration. 
Finally, all input images are center-cropped and resized to $448\times448$ before being fed into the model.
\paragraph{Evaluation Dataset}
We evaluate our simultaneous geometry and semantic prediction on two widely used multi-view datasets: ScanNet~\cite{dai2017scannet} and the DL3DV-10K~\cite{ling2024dl3dv}. 
Following AnySplat~\cite{jiang2025anysplat}, we first sample 72 views from the original video sequence based on spatial distribution, and further downsample them to 56 and 32 views. With the test interval set to 8, as in 3DGS~\cite{kerbl3Dgaussians}, the corresponding numbers of context views become 32, 48, and 64, respectively. For the sparse-view setting, we use a test interval of 1.
For datasets with semantic annotations (e.g., ScanNet\cite{dai2017scannet}), we map the thousands of different labels into a set of common labels following~\cite{fan2024largespatialmodelendtoend}. To evaluate our model under more challenging and unconstrained scenarios, we additionally test on DL3DV-10K~\cite{ling2024dl3dv}, which contains unbounded scenes, diverse environments, and varying lighting conditions. Since DL3DV does not provide semantic annotations, we follow Feature-3DGS~\cite{zhou2024feature} and adopt semantic masks predicted by LSeg~\cite{lseg} as pseudo ground truth. This allows us to evaluate how effectively our method lifts inherently inconsistent 2D semantic features into a geometrically consistent 3D representation.

\section{Additional Experiments and Results}
\label{sec:app_exp}
\paragraph{Additional Ablation Studies}
In Tab. \ref{tab:ablation_distill}. The results demonstrate that the distilled geometry is critical; without it, the model suffers from overfitting to the context views and fails to generalize to novel views. This highlights that our distillation mechanism is a necessary component for enabling robust, scalable reconstruction without human labels.
\begin{table}[t]
\centering
\caption{Ablation for Geometric Distillation.}
\label{tab:ablation_distill}
\resizebox{0.8\linewidth}{!}{
\begin{tabular}{lccc}
\toprule
Method & PSNR$\uparrow$ & SSIM$\uparrow$ & LPIPS$\downarrow$ \\
\midrule
Ours w/o distill loss & 9.37 & 0.248 & 0.783 \\
Ours w/ distill loss &  17.85 & 0.699 & 0.380 \\
\bottomrule
\label{tab:rebuttal_ablation}
\end{tabular}
}
\end{table}

\paragraph{Additional Baseline Comparisons}
\begin{table}[t!]
\caption{Quantitative Results on Novel View Synthesis and Semantic Segmentation on ScanNet.}
\centering
\renewcommand{\arraystretch}{1.15}
\setlength{\tabcolsep}{6pt}
\resizebox{0.9\linewidth}{!}{
\begin{tabular}{l|cccccc|cccccc}
\toprule
 & \multicolumn{6}{c|}{2 Views} & \multicolumn{6}{c}{16 Views} \\
\cmidrule(lr){2-7} \cmidrule(lr){8-13}
 & PSNR$\uparrow$ & SSIM$\uparrow$ & LPIPS$\downarrow$ & mIoU$\uparrow$ & Acc.$\uparrow$ & Time(s)$\downarrow$ 
 & PSNR$\uparrow$ & SSIM$\uparrow$ & LPIPS$\downarrow$ & mIoU$\uparrow$ & Acc.$\uparrow$ & Time(s)$\downarrow$ \\
\midrule
Uni3R
& {15.57} & {0.649} & {0.448} & {0.354} & {0.716} & {0.626s} 
& {17.98} & {0.737} & {0.402} & {0.437} & {0.769} & {0.928s} \\
Ours 
& {22.70} & {0.787} & {0.285} & {0.486} & {0.754} & {0.884s} 
& {22.19} & {0.769} & {0.295} & {0.500} & {0.751} & {1.2s} \\
\bottomrule
\end{tabular}}
\label{tab:comparsion_uni3r}
\end{table}

\begin{figure}[htbp]
    \centering
    \includegraphics[width=1.0\linewidth]{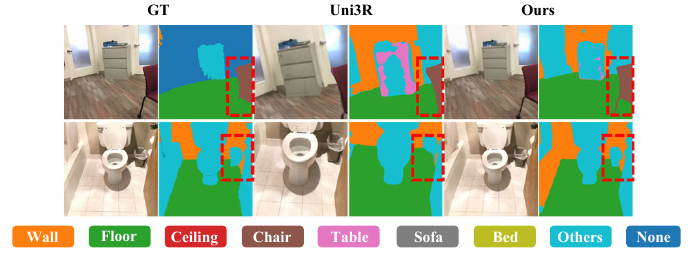}
    \vspace{-0.3cm}
    \caption{Qualitative Results on Novel View Synthesis and Semantic Segmentation on ScanNet.}

    \label{fig:comparsion_uni3r}
    % \vspace{-0.3cm}
\end{figure}
We further evaluate our method against the state-of-the-art approach Uni3R~\cite{sun2025uni3runified3dreconstruction}. As shown in Tab.~\ref{tab:comparsion_uni3r} and Fig.~\ref{fig:comparsion_uni3r}, our method consistently outperforms this baseline.

\begin{figure*}[htbp]
    \centering
    \includegraphics[width=.9\linewidth]{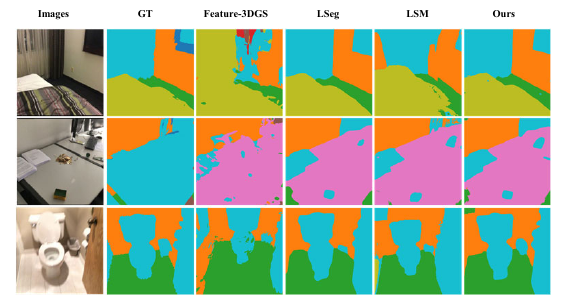}
    \vspace{-0.3cm}
    \caption{Qualitative results of open-volcabulory semantic segmentation.}

    \label{fig:app_seg}
    % \vspace{-0.3cm}
\end{figure*}

\begin{figure*}[htbp]
    \centering
    \includegraphics[width=.7\linewidth]{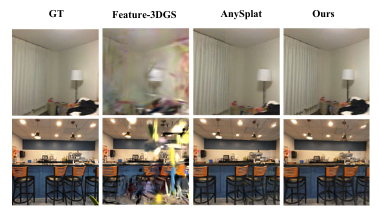}
    \vspace{-0.3cm}
    \caption{Qualitative results of novel view synthesis.}

    \label{fig:app_nvs}
    % \vspace{-0.3cm}
\end{figure*}

\begin{figure*}[htbp]
    \centering
    \includegraphics[width=.9\linewidth]{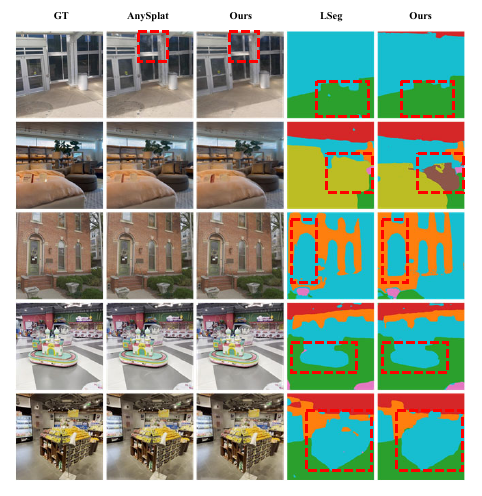}
    \vspace{-0.3cm}
    \caption{Qualitative results on DL3DV-10K~\cite{ling2024dl3dv} demonstrating generalization across diverse indoor and outdoor scenes.}

    \label{fig:app_nvs_seg}
    % \vspace{-0.3cm}
\end{figure*}

\paragraph{Additional Qualitative Results} We provide additional qualitative results of our model on simultaneous geometry and semantic reasoning in ScanNet~\cite{dai2017scannet} and DL3DV-10K~\cite{ling2024dl3dv} in Figs.~\ref{fig:app_seg}, \ref{fig:app_nvs}, and \ref{fig:app_nvs_seg}. As shown in Fig.~\ref{fig:app_seg}, our method preserves sharp boundaries between different semantic regions. Even when the 2D semantic features are inconsistent, the proposed Geometry-Guided Feature Warping effectively injects 3D awareness into the semantic features, resulting in improved generalization across challenging viewpoints.

Moreover, with the Semantic-Aware Voxelization module, our model reduces local visual artifacts by enforcing semantic consistency within each voxel, as illustrated in Fig.~\ref{fig:app_nvs}. Finally, benefiting from our fully annotation-free training strategy, the model requires no explicit semantic annotations and can be trained on large, diverse datasets such as DL3DV-10K~\cite{ling2024dl3dv}. This enables strong generalization across indoor and outdoor scenes under varying lighting conditions, as demonstrated in Fig.~\ref{fig:app_nvs_seg}.
\vspace{-0.1cm}

\end{document}